\documentclass[11pt]{article}

\usepackage[final]{acl}

\usepackage{booktabs}
\usepackage{times}
\usepackage{latexsym}
\usepackage[T1]{fontenc}
\usepackage[utf8]{inputenc}
\usepackage{microtype}
\usepackage{inconsolata}
\usepackage{graphicx}
\usepackage{booktabs}
\usepackage{multirow}
\usepackage{amsmath}
\usepackage{float}
\usepackage{array}

\title{When Better Codebooks Are Not Enough: Predictive Performance and Rule Adherence in LLM Political Event Coding}

\author{
Zixian He\thanks{This research was conducted during Zixian He's remote research internship under the supervision of Yibo Hu.} \\
Independent Researcher \\
\texttt{zixianh@usc.edu}
\And
Bharath Raahul Murugesan \\
Illinois Institute of Technology \\
\texttt{bmurugesan@hawk.illinoistech.edu}
\AND
Patrick T.~Brandt \\
The University of Texas at Dallas \\
\texttt{pbrandt@utdallas.edu}
\And
Yibo Hu\thanks{Corresponding author.} \\
Illinois Institute of Technology \\
\texttt{yhu89@illinoistech.edu}
}

\begin{document}

\maketitle

\begin{abstract}
High accuracy alone does not establish whether an LLM follows a codebook. This matters because many social-science studies rely on expert-written codebooks to turn text into structured data. We study this problem in political event coding, a source–target relation classification task in which a model identifies the action that one actor directs toward another under detailed coding rules.
We test whether expert codebooks become more effective when presented to LLMs through clearer definitions, examples, retrieved context, and rules for difficult cases. We then examine how model predictions respond to controlled changes in label names, codebook order, and label–definition mappings.
Clearer codebooks improve fine-grained classification: mean root-level macro-F1 rises from 0.457 with label names alone to 0.633 with enriched guidance across four open-source LLMs. Yet reversing the order of entries in the enriched codebook changes 21.2\% of predictions on a 312-example subset, even though the labels and definitions remain unchanged. When label names are reassigned to different definitions, performance is also low.
These findings highlight the need to evaluate both predictive performance and responses to codebook changes when LLM-generated labels are used as data in social-science research. \footnote{Code, prompts, and codebooks are available at \url{https://github.com/yibo-hu-lab/event-coding-reliability}}
\end{abstract}

\section{Introduction}
\label{sec:introduction}

\begin{figure}[t]
    \centering
    \includegraphics[width=\columnwidth]{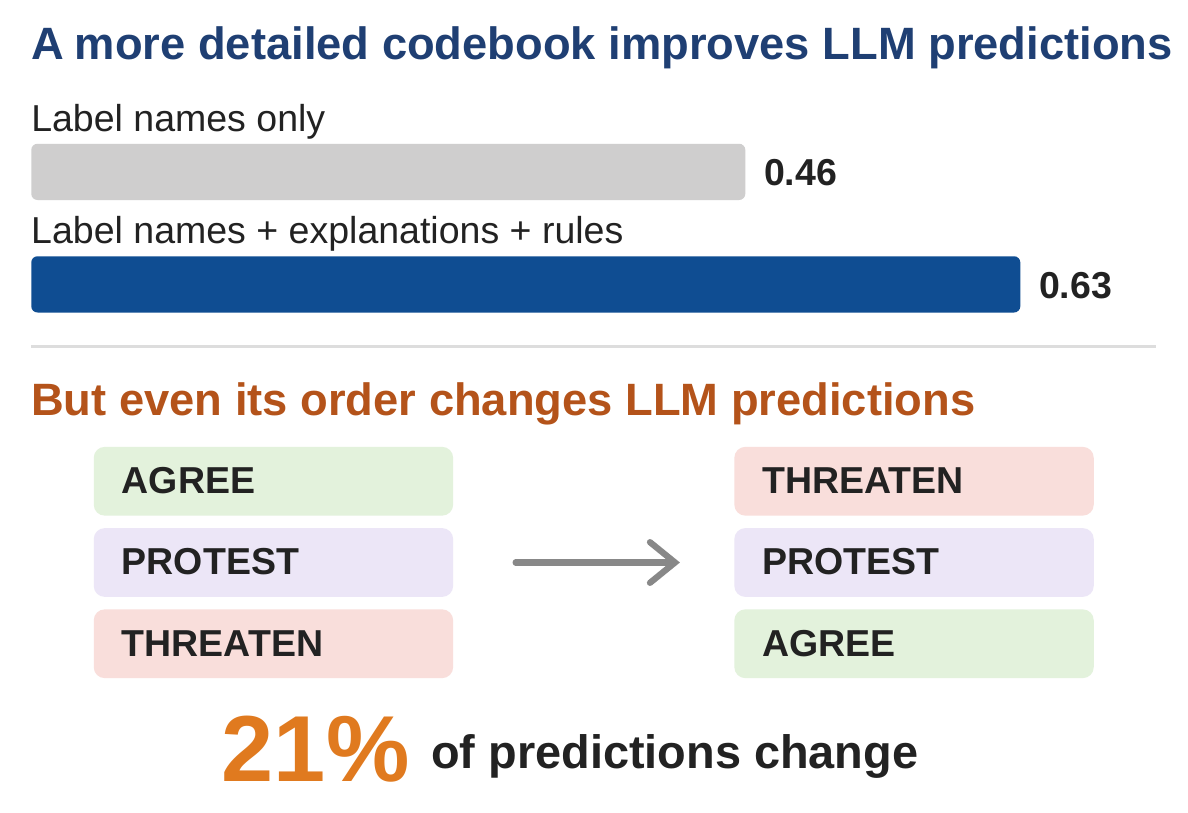}
    \caption{\textbf{Better codebooks, inconsistent answers.} Adding explanations and coding rules to category names improves mean macro-F1. Yet reversing the order of the same codebook entries changes model predictions, even though the categories and their definitions remain unchanged.}
    \label{fig:teaser}
\end{figure}

Many social-science studies rely on political event coding to turn news reports into structured records of who did what to whom \citep{raleigh2010acled,hu2024leveraging}. These records capture actions such as cooperation, protests, threats, and violence, allowing researchers to monitor, analyze, and forecast political conflict and mediation processes \citep{schrodt1996using,schrodt2003evaluating,schrodt2004using,brandt2011,brandt2014forecasting}.

Expert-written codebooks specify how these actions should be classified, including category boundaries and rules for ambiguous cases \citep{krippendorff2019content,neuendorf2017content}. Codebooks are essential measurement resources, but they are costly to build and apply because they require domain expertise, annotator training, and reliability checks.

Prior event-coding systems have relied on event ontologies, dictionaries, and pattern-based knowledge bases to turn news text into structured political event records \citep{mcclelland2006weis,azar1980copdab,bond2003idea}. Later systems and resources further formalized automated political event coding \citep{gerner2002,boschee2015icews,plover2018,lu2017universal}. These resources make event coding reusable and interpretable, but static dictionaries and pattern-matching systems can be brittle when language use, domains, or coding schemes change. More recent supervised and pretrained language-model approaches improve flexibility for socio-political event extraction and classification, but they often require annotated data and may need relabeling when ontologies evolve \citep{buyukoz2020analyzing,hu2022conflibert,parolin2021come,parolin2021eventcoding,parolin2022multicoped,parolin2022conflit5,hu2024leveraging}.

We focus on one part of political event coding: given a news report and a specified source--target actor pair, the model predicts the action directed from the source to the target. Ontologies such as CAMEO (Conflict and Mediation Event Observations) and PLOVER (Political Language Ontology for Verifiable Event Records) use hierarchical labels and event-mode rules that can assign different labels to similar wording depending on whether an action occurred, was promised, refused, negated, or halted \citep{gerner2002,hu2024leveraging}. These distinctions make political event coding a demanding test of whether LLMs follow codebook rules rather than matching familiar label semantics.

The cost of applying codebooks by hand motivates the use of LLMs to apply them directly. Recent work shows that definitions, examples, and coding instructions can improve zero-shot classification, especially when labels are specialized or difficult to infer from names alone \citep{hu2024leveraging,ruckdeschel2025,stuhler2025promptbooks}. Yet strong predictive performance does not necessarily mean that models are truly applying the codebook definitions and rules. This concern aligns with broader questions about the reliability of LLM evaluation \citep{aiyappa2023trust}. Models may instead rely on familiar label semantics, prompt order, or pretrained associations rather than the operational logic defined by the codebook \citep{halterman2025codebook_llms}.

Figure~\ref{fig:teaser} illustrates this problem: richer codebooks improve prediction, yet reversing the order of their entries changes about one in five answers.

We study two questions:
\begin{enumerate}
    \item  Can richer codebook guidance improve predictive performance?
    \item Does stronger predictive performance necessarily imply adherence to the coding rules?
\end{enumerate}

To answer them, we evaluate event-coding LLMs in two ways: predictive performance under the original codebook and responses to controlled changes in the codebook. This design separates whether a model predicts the expected label from whether its predictions remain tied to the supplied coding rules.

Our contributions are threefold:
(1) we study codebook-guided LLM evaluation in a demanding political event-coding setting with hierarchical labels and event-mode distinctions;
(2) we show that richer codebook guidance substantially improves fine-grained event classification across open-source LLMs; and
(3) we adapt controlled perturbation probes to source--target event coding, testing sensitivity to codebook order, generic label names, and swapped label--definition mappings.

\section{Preliminaries}
\label{sec:preliminaries}

\subsection{Codebooks as Measurement Systems}
\label{sec:codebooks-measurement}

Text-based social-science measurement often requires turning documents into structured variables. In this setting, codebooks specify category definitions and label boundaries that guide how qualitative evidence is converted into analyzable observations \citep{krippendorff2019content,neuendorf2017content}. They therefore provide the measurement rules that coders, including models used as coders, are expected to follow.

The same structure that makes codebooks useful also makes them costly to build and apply. Codebook construction requires domain expertise and iterative refinement, while consistent use depends on annotator training and reliability assessment. %
These costs motivate the use of large language models as potential codebook users, without training a separate supervised model for each coding scheme. Political event coding provides a useful testbed because its labels depend on fine-grained coding rules rather than keyword matching alone.

\subsection{Political Event Coding as a Challenging Testbed}
\label{sec:event-coding-testbed}

Political event coding converts news text into structured records of political interactions. Full event coding may involve identifying actors, actions, and relations; this paper focuses on the relation-classification component, in which the model assigns the political relation that a specified source actor directs toward a specified target actor.

For this study, we use PLOVER, a CAMEO-based ontology designed for source–target political relation classification \citep{gerner2002,plover2018,hu2024leveraging}. PLOVER labels are evaluated at three levels: binary cooperation versus conflict (2 classes), quad-level verbal or material cooperation and conflict (4 classes), and root-level categories (15 classes) such as agreement, aid, rejection, threat, protest, coercion, and assault.

\begin{figure}[t]
\centering
\fbox{%
\begin{minipage}{0.95\columnwidth}
\footnotesize
\raggedright
\textbf{Event-mode distinction in protest-related coding}

\vspace{0.4em}
\textbf{Source:} Salaheddine Sidhoum \\
\textbf{Target:} Algerian security forces

\vspace{0.45em}
\textit{Announced protest action}\\
\textbf{Text:} ``Salaheddine Sidhoum announced that he would begin a
hunger strike against abuses by Algerian security forces.''\\
\textbf{Binary:} Conflict \\
\textbf{Quad:} Verbal Conflict \\
\textbf{Root:} THREATEN

\vspace{0.65em}
\textit{Completed protest action}\\
\textbf{Text:} ``Salaheddine Sidhoum staged a hunger strike in protest
against abuses by Algerian security forces.''\\
\textbf{Binary:} Conflict \\
\textbf{Quad:} Material Conflict \\
\textbf{Root:} PROTEST

\vspace{0.65em}
\textit{Ended protest action}\\
\textbf{Text:} ``Salaheddine Sidhoum ended his hunger strike after
negotiations with Algerian security forces.''\\
\textbf{Binary:} Cooperation \\
\textbf{Quad:} Material Cooperation \\
\textbf{Root:} YIELD
\end{minipage}%
}
\caption{Illustration of event-mode distinctions in PLOVER. The same ordered source--target pair receives different labels depending on whether the protest action is announced, carried out, or ended. The completed case shortens the news example used in Section~\ref{sec:task-setup}; the announced and ended versions are constructed for illustration.}
\label{fig:plover-event-mode-example}
\end{figure}

As Figure~\ref{fig:plover-event-mode-example} shows, the same ordered source--target pair is coded THREATEN, PROTEST, or YIELD depending on whether the protest action is announced, carried out, or ended. Correct coding therefore requires the model to preserve actor direction while distinguishing event mode. Such cases require applying the codebook rules rather than matching keywords to familiar labels.

\begin{figure*}[t]
    \centering
    \includegraphics[width=\textwidth]{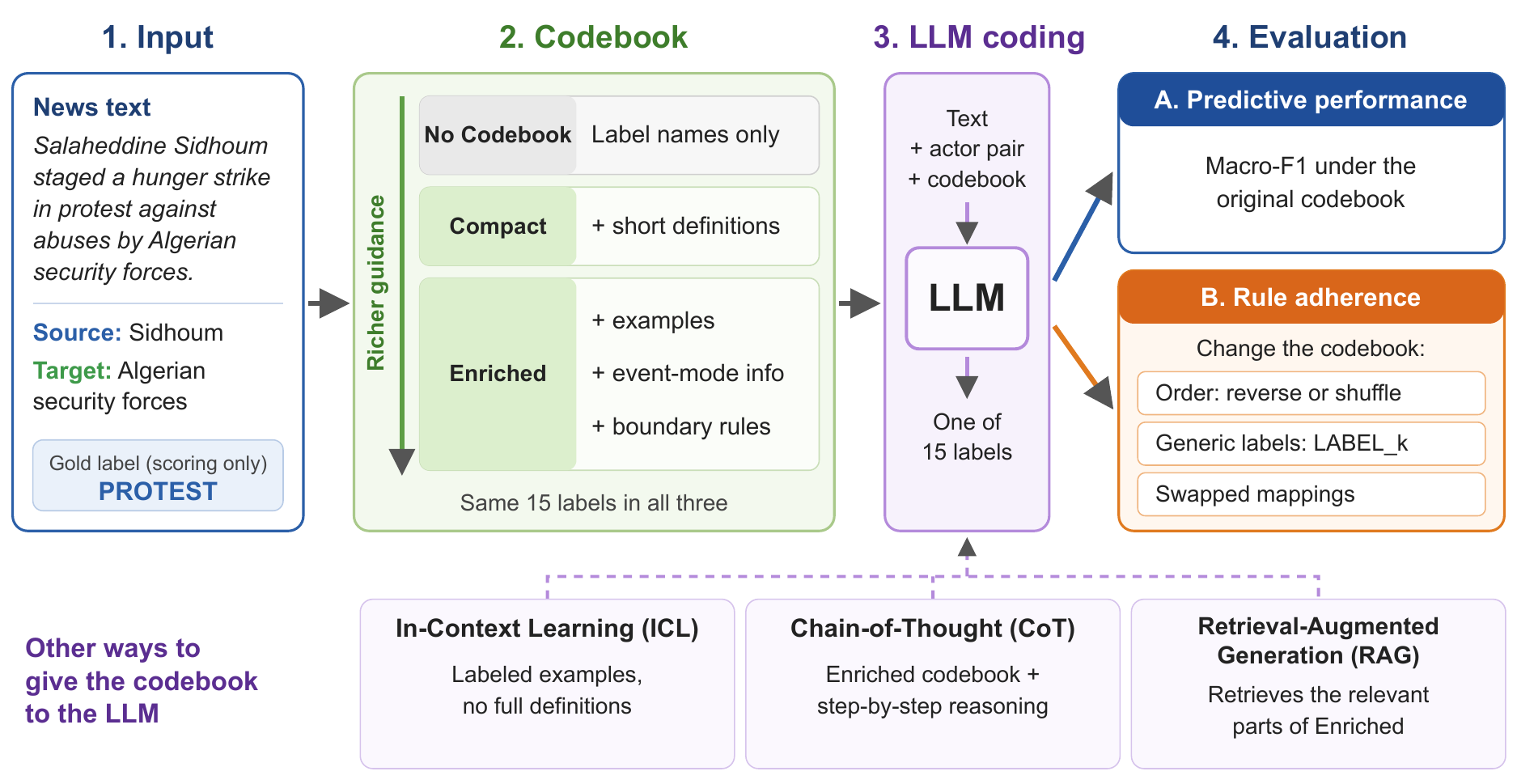}
    \caption{\textbf{Overview of the experimental framework for political event coding.} We compare different amounts and forms of coding guidance, then evaluate both predictive performance and how model outputs respond to changes in the codebook.}
    \label{fig:framework}
\end{figure*}

\begin{table*}[t]
\centering
\caption{Representative differences between the Compact and Enriched codebooks for three pairs of neighboring labels. Both codebooks use the same 15 root labels.}
\label{tab:cb1-cb2-summary}
\small
\setlength{\tabcolsep}{4pt}
\renewcommand{\arraystretch}{1.15}
\begin{tabular}{@{}
>{\raggedright\arraybackslash}p{0.16\textwidth}
>{\raggedright\arraybackslash}p{0.27\textwidth}
>{\raggedright\arraybackslash}p{0.30\textwidth}
>{\raggedright\arraybackslash}p{0.21\textwidth}@{}}
\toprule
Boundary case
& Compact guidance
& Enriched addition
& Coding implication \\
\midrule
Future vs completed aid
& AGREE covers promises and willingness to cooperate, including cooperation reported in future tense; AID covers provisions of material aid.
& Adds examples such as ``promised aid'' and ``will provide assistance'' under AGREE and ``sent humanitarian supplies'' under AID.
& Separates stated commitment from completed material cooperation. \\
\addlinespace
Aid vs mobilization
& AID covers material aid, whereas MOBILIZE covers military or police moves short of force.
& Adds a rule that peacekeeping forces, workers, or observers sent to help are coded as AID, not MOBILIZE.
& Separates assistance from military force posture. \\
\addlinespace
Demand vs protest
& REQUEST covers verbal demands; demands made as demonstrations are coded as PROTEST.
& Adds protest-form examples such as ``marched in protest'' and ``organized a strike.''
& Prevents verbal demand wording from overriding the protest event. \\
\bottomrule
\end{tabular}
\end{table*}

\subsection{Codebook Guidance and Related Work}
\label{sec:codebook-operationalization-reliability}

Prior work has explored different ways to incorporate coding knowledge into automated classification. Zero-shot and natural language inference (NLI) methods use label descriptions, hypotheses, or ontology information to classify text \citep{bowman2015snli,obamuyide2018zero,yin2019benchmarking,huang2018zeroshot,geng2021ontozsl}. Within political event coding, \citet{hu2024leveraging} show that PLOVER codebook knowledge can be converted into NLI hypotheses for zero-shot relation classification. Their strongest system relies on hand-written hypotheses and a task-specific decision tree. 
Prompt- and promptbook-based approaches similarly place definitions, examples, and coding instructions directly into model inputs \citep{ruckdeschel2025,stuhler2025promptbooks,than2025future}. These approaches reduce retraining or relabeling when schemas change, but much of this work focuses on whether models predict the correct label under the original codebook.

Accurate outputs do not by themselves show that an LLM is reliable. LLM outputs can shift with context that should not affect the answer \citep{qu2026mislead,hu2026speakerfree,guerra2026judge,hu2026socialpressure}, and models have limited insight into their own errors \citep{hu2026oneaxis}. Work on reasoning traces shows that a correct answer can rest on flawed reasoning \citep{molfese2026retraceqa}. In codebook-based measurement, \citet{halterman2025codebook_llms} evaluate LLMs as measurement tools for political science concepts, including whether their outputs respond to the supplied codebook.

We bring this question to directed, hierarchical event coding, where each prediction is conditioned on a marked actor pair, event mode, and category-boundary rules. Unlike the NLI-based approach of \citet{hu2024leveraging}, we provide adapted codebook text directly in the prompt of a general-purpose LLM. Richer codebook guidance may improve predictive performance by making coding rules easier to access, but performance under a single prompt does not establish adherence to those rules. We therefore test whether predictions remain tied to the intended coding rules under controlled codebook changes.

\section{Approach}
\label{sec:approach}

This section defines the source--target classification task, describes how codebook knowledge is represented for LLMs, and introduces probes of rule adherence. Figure~\ref{fig:framework} summarizes the study design: each input is a news text with a marked source and target; the model uses different levels of codebook guidance or alternative prompting and retrieval methods; and its predictions are evaluated both under the original codebook and under controlled changes to the codebook.

\subsection{Task Setup}
\label{sec:task-setup}

We study source--target political relation classification: given a text and a specified source--target actor pair, the model predicts the event label for the directed relation. We evaluate this task under the PLOVER hierarchy introduced above, with root-level coding serving as the strongest test of codebook boundary-rule application.

We use one PLV example throughout: ``Salaheddine Sidhoum, Algeria's leading human rights activist, staged a 24-hour hunger strike in prison in protest against widespread human rights violations by Algerian security forces.'' Salaheddine Sidhoum is the marked source, Algerian security forces are the marked target, and the gold label is PROTEST, as in the completed case of Figure~\ref{fig:plover-event-mode-example}.

\subsection{Codebook Guidance Strategies}
\label{sec:codebook-operationalization-strategies}

We compare several ways of presenting codebook knowledge to the model, in three groups.

\paragraph{Codebook content.}
The No Codebook baseline provides only the text, the marked source--target pair, and the valid label set. It tests what can be inferred from label names and pretrained knowledge alone. \textbf{Compact} codebook prompting adds short natural-language definitions and assigns each label to one of four broad event groupings (the quad level). \textbf{Enriched} codebook prompting keeps the same label inventory as Compact but adds worked examples, event-mode guidance, boundary notes, and disambiguation rules for commonly confused categories. These additions clarify how the original codebook should be applied without changing the underlying label space.

Table~\ref{tab:cb1-cb2-summary} shows what Enriched adds for three pairs of neighboring labels. Because Enriched combines multiple forms of guidance, we interpret Compact--Enriched differences as the effect of the enriched codebook package rather than as an ablation of individual components.

\paragraph{Alternative prompting and retrieval.}
In-context learning (\textbf{ICL}) provides labeled demonstrations without full codebook definitions \citep{brown2020language}.
Chain-of-thought (\textbf{CoT}) prompting uses the Enriched codebook with an added step-by-step reasoning instruction before the final label prediction \citep{wei2022chain,suzgun2023}.
Retrieval-augmented generation (\textbf{RAG}) retrieves input-relevant definitions, examples, or rules from the Enriched codebook instead of presenting the full codebook \citep{lewis2020retrieval}. Implementation details are provided in Appendix~\ref{app:rag-details}.

\paragraph{Prior event-coding systems.}
We also report two prior baselines from \citet{hu2024leveraging}: Universal PETRARCH (\textbf{UP}) \citep{lu2017universal}, a traditional dictionary- and pattern-based event coder, and \textbf{ZSP Tree}, a hierarchical zero-shot system that scores codebook-derived NLI hypotheses and resolves event mode through a multi-level decision tree.

\subsection{Testing Adherence to Coding Rules}
\label{sec:behavioral-reliability-probes}

Rule adherence refers to whether predictions follow the supplied coding definitions, independent of label names or entry order. We test adherence to the coding rules through controlled codebook perturbations adapted to source--target event coding. Each probe holds the input text and source--target actor pair fixed while changing the codebook presentation. The goal is to test whether predictions remain tied to the intended coding rules rather than relying on label names, entry order, or pretrained semantic associations.

\paragraph{Order probe.}
We reverse or shuffle the codebook entries for Compact, Enriched, and CoT; for ICL, only the order of the allowed labels changes. Label names and definitions are unchanged, so predictions should remain stable. In the running example, the expected output stays PROTEST.

\paragraph{Generic-label probe.}
We replace each meaningful label name with a neutral placeholder such as \texttt{LABEL\_1} or \texttt{LABEL\_2}, assigned by the fixed codebook order, while leaving the associated definitions unchanged. Gold labels are remapped with the same assignment, so the expected output is the placeholder attached to the matching definition; in the running example, it is the placeholder that replaces PROTEST.

\paragraph{Swapped-mapping probe.}
We keep the definitions fixed and reassign the label names to them, so that no label keeps its original definition. Gold labels are remapped with the same permutation, so the expected output is the name now attached to the matching definition. In the running example, the protest definition is renamed MOBILIZE, so the expected output becomes MOBILIZE.

The same shuffle and reassignment are used in all examples and models. These probes are not intended to simulate realistic ontology revisions directly. Instead, they test whether predictions follow the supplied codebook when its presentation or label--definition mapping changes.

The generic-label and swapped-mapping probes rename labels but do not update every reference to the original names: the instruction and some definitions still name original labels, and the swapped-mapping probe also keeps the original output examples (Appendix~\ref{app:prompting-conditions}). The results of these probes do not isolate the effect of familiar label meanings.

\subsection{Evaluation Metrics}
\label{sec:behavioral-metrics}

Predictive performance is measured with macro-F1 on held-out test examples. For PLV, macro-F1 is computed over the 15 root-level PLOVER labels. For AW, macro-F1 is computed over the binary Cooperation/Conflict labels. For PLV, outputs that cannot be parsed into a valid label are counted as incorrect rather than dropped; for AW, they are assigned to Conflict.

Each probe is scored according to the behavior it tests. Original-condition accuracy measures the share of predictions matching the gold label under the unmodified codebook. For the order probe, Fleiss' $\kappa$ measures agreement among predictions under the original, reversed, and shuffled codebook orders ($\kappa=1$ indicates identical predictions), and we also report the share of predictions that change. Generic-label and swapped-mapping probes are evaluated with weighted F1 against the remapped gold labels.

\section{Experiments}
\label{sec:experiments}

\begin{table*}[t]
\centering
\caption{Predictive performance across codebook-content conditions. Scores are mean macro-F1 across four open-source instruction-tuned LLMs. PLV is evaluated at the root level and AW at the binary level. $\Delta$ is the difference from No Codebook. Per-model PLV results are in Appendix~\ref{app:cross-model-plv}.}
\label{tab:predictive-results}
\small
\setlength{\tabcolsep}{6pt}
\renewcommand{\arraystretch}{1.15}
\begin{tabular}{lp{0.48\textwidth}cccc}
\toprule
& & \multicolumn{2}{c}{PLV Root} & \multicolumn{2}{c}{AW Binary} \\
\cmidrule(lr){3-4}
\cmidrule(lr){5-6}
Method & Codebook content & F1 & $\Delta$ & F1 & $\Delta$ \\
\midrule
No Codebook
& Label names only
& 0.457 & -- & 0.757 & -- \\
Compact
& Label names, short definitions, and quad-level groupings
& 0.574 & +0.117 & 0.782 & +0.025 \\
Enriched
& Compact plus examples, event-mode guidance, boundary notes, and disambiguation rules
& \textbf{0.633} & +0.176 & \textbf{0.794} & +0.037 \\
\bottomrule
\end{tabular}
\end{table*}

\subsection{Dataset and Setup}

\label{sec:datasets-setup}

Our main evaluation uses the PLOVER-based PLV benchmark for source--target political relation classification \citep{hu2024leveraging}. It contains 1,033 annotated source--action--target examples, each with an input text, a marked source actor, a marked target actor, and a directed event label. We report macro-F1 at three granularities: binary Cooperation/Conflict, quad-level verbal/material cooperation or conflict, and root-level categories from AGREE to ASSAULT.

We also evaluate on AW (ACE/WikiEvents), a cross-domain binary Cooperation/Conflict task derived from the ACE event extraction corpus \cite{ace04}  and WikiEvents \cite{li2021document} by mapping seven NLP event types to PLOVER categories \citep{hu2024leveraging}; we evaluate the 802 of its 805 test examples whose event types fall in these seven types. Because the mapping between NLP event types and PLOVER Rootcodes is approximate, only binary evaluation is reliable on this dataset.

For PLV, root-level predictions are also mapped upward through the PLOVER hierarchy. The probes are run on evaluation subsets of 312 PLV and 253 AW examples by rerunning the same inputs under controlled codebook perturbations: reordered entries, generic label names, and swapped label--definition mappings. PLV probes use root labels, and AW probes use binary labels.

The main predictive comparison focuses on three LLM-facing conditions: No Codebook, Compact, and Enriched. The archived runs use greedy decoding rather than stochastic sampling. We report results across four open-source instruction-tuned models: Gemma2:9B, Qwen2.5:7B, Mistral-7B-Instruct, and Llama-3.1-8B-Instruct. The Compact and Enriched codebooks, the ICL demonstrations, and the RAG example bank are derived from the PLOVER/CAMEO coding manual and its worked examples. ICL, CoT, and RAG are reported as alternative prompting and retrieval methods, and UP and ZSP Tree as prior baselines (Section~\ref{sec:codebook-operationalization-strategies}). Additional ZSP, RAG, and prompting details are reported in Appendices~\ref{app:prompts}--\ref{app:rag-details}.

\subsection{Codebook Guidance Improves Prediction}
\label{sec:predictive-results}

Table~\ref{tab:predictive-results} reports the main predictive results on PLV and AW. On PLV, mean root-level macro-F1 rises from 0.457 with No Codebook to 0.574 with Compact and 0.633 with Enriched, and Enriched is highest for all four models. The gain is small at the binary level, where mean PLV macro-F1 rises from 0.893 to 0.914, and large at the finer levels, from 0.597 to 0.808 at the quad level and from 0.457 to 0.633 at the root level. At these levels, models must distinguish neighboring event categories rather than only broad conflict/cooperation polarity. AW shows the same broad direction, but the gains are smaller and less uniform because the task is binary.

\begin{table}[t]
\centering
\caption{Alternative prompting and retrieval methods and prior baselines. PLV reports root-level macro-F1, AW binary macro-F1. ICL, CoT, and RAG are means across the four open-source LLMs; UP and ZSP Tree are single prior systems.}
\label{tab:alternative-variants}
\footnotesize
\setlength{\tabcolsep}{4pt}
\begin{tabular*}{\columnwidth}{@{\extracolsep{\fill}}p{0.14\columnwidth}p{0.48\columnwidth}rr}
\toprule
Method & Codebook use & PLV & AW \\
\midrule
ICL & Labeled demonstrations without full definitions & 0.490 & 0.790 \\
CoT & Enriched codebook with step-by-step reasoning & 0.477 & 0.787 \\
RAG & Retrieved content from the Enriched codebook & 0.588 & 0.813 \\
\midrule
UP & Dictionary and pattern matching & 0.463 & 0.672 \\
ZSP Tree & Codebook-derived NLI hypotheses with a decision tree & 0.824 & 0.880 \\
\bottomrule
\end{tabular*}
\end{table}

The prior baselines provide context for the LLM results (Table~\ref{tab:alternative-variants}). UP, a classic dictionary-based event coder, performs well below Compact, Enriched, and RAG, while ZSP Tree remains the strongest baseline. This suggests that simple codebook prompting improves over static dictionary matching, but still falls short of specialized NLI-based event-mode disambiguation: ZSP Tree exceeds the best LLM setting by 0.19 root-level macro-F1 (0.824 versus 0.633). Among the alternative prompting and retrieval methods, RAG is strongest on PLV but remains below Enriched prompting, while it slightly exceeds Enriched on AW. CoT does not outperform Compact or Enriched on PLV.

The PLV--AW contrast further suggests that codebook guidance matters most when labels require boundary-rule application. In AW, many cases can be resolved from broad Cooperation/Conflict polarity. In PLV root coding, however, the model must choose among neighboring categories that share surface-level wording but differ in polarity or mode, such as AGREE versus REQUEST, or AID versus SANCTION. These distinctions depend more directly on event-mode and boundary rules, which helps explain why Enriched's additional gain over Compact is largest at the root level. Full binary and quad-level cross-model results are reported in Appendix~\ref{app:cross-model-plv}.

To add uncertainty estimates for the main PLV codebook comparisons, we compute nonparametric bootstrap confidence intervals for root-level macro-F1 differences \citep{efron1979bootstrap}. The intervals support the broad benefit of adding compact codebook guidance: Compact improves over the no-codebook baseline for all four models, by 0.075 to 0.146. The additional Enriched-over-Compact gain is smaller, from 0.031 to 0.082, and positive for all four models; for Qwen2.5, the lower bound of its interval is close to zero and rounds to 0.000. Table~\ref{tab:bootstrap-plv-delta} in Appendix~\ref{app:bootstrap} reports the intervals.

\subsection{Codebook Changes Reveal Gaps in Rule Adherence}
\label{sec:behavioral-reliability-results}

We next examine how predictions change when the codebook is reordered or labels are renamed. Table~\ref{tab:probe-breakdown-main} reports the probe results on PLV, averaged across four instruction-tuned models; AW results are in Appendix~\ref{app:full-behavioral-probes}. The table separates direct Compact and Enriched prompting from prompting and retrieval variants, so the Compact--Enriched comparison is visible rather than averaged away.

\begin{table}[t]
\centering
\caption{Probe-level PLV results (root labels, 312 examples), averaged across four open-source models. Original is accuracy under the unmodified codebook; order agreement is Fleiss' $\kappa$ across original, reversed, and shuffled orders; the last two columns are weighted F1 against the remapped gold labels. Values are diagnostic and should be read by probe type rather than as a single leaderboard. Order agreement is not defined for RAG, which retrieves local chunks rather than an ordered codebook. ICL uses no codebook definitions; CoT and RAG use the Enriched codebook.}
\label{tab:probe-breakdown-main}
\footnotesize
\setlength{\tabcolsep}{3pt}
\begin{tabular*}{\columnwidth}{@{\extracolsep{\fill}}lcccc}
\toprule
Method
& \shortstack{Original\\accuracy}
& \shortstack{Order\\agreement}
& \shortstack{Generic-\\label F1}
& \shortstack{Swapped-\\mapping F1} \\
\midrule
Compact  & 0.612 & 0.743 & 0.566 & 0.105 \\
Enriched & 0.619 & 0.782 & 0.519 & 0.081 \\
\midrule
ICL      & 0.454 & 0.668 & 0.152 & 0.064 \\
CoT      & 0.642 & 0.745 & 0.625 & 0.152 \\
RAG      & 0.633 & N/A   & 0.522 & 0.191 \\
\bottomrule
\end{tabular*}
\end{table}

Order perturbations are not harmless: reversing the codebook changes 25.5\% of Compact predictions and 21.2\% of Enriched predictions (Appendix~\ref{app:full-behavioral-probes}). For Enriched, accuracy is similar before and after reversal (0.619 versus 0.625), despite these changes in individual predictions. Under generic labels, Compact, Enriched, CoT, and RAG retain substantially higher scores than ICL, which lacks the full definitions. Swapped-mapping F1 ranges from 0.064 to 0.191.

The Compact--Enriched contrast clarifies the main pattern. Enriched improves order agreement relative to Compact, but it does not improve generic-label or swapped-mapping F1. This supports the interpretation that enrichment helps with the original ontology and some presentation changes, but not with the probes that rename labels.

\subsection{Analysis}
\label{sec:analysis}

Taken together, the results show that richer codebook guidance improves hard event coding, but high predictive accuracy does not necessarily imply adherence to the coding rules. Enriched is strongest in the main predictive comparison and also the most stable under reordering.

Models can perform well under the original ontology while remaining sensitive to how the codebook is presented.
Under the swapped-mapping prompt, CoT outputs frequently retain the original label.

For measurement applications, this distinction is consequential. Event labels are often treated as observed data in downstream analyses, but if a model changes labels under codebook-preserving perturbations, the resulting variables may reflect prompt artifacts rather than the intended political concepts. Failures to follow the coding rules at the coding stage therefore become measurement uncertainty and can threaten measurement validity \citep{adcock2001measurement}. Codebook-guided LLMs should therefore be evaluated not only by predictive performance, but also by whether they preserve the coding rules specified by the codebook under controlled perturbations.

\section{Conclusion}
\label{sec:conclusion}

This work studies codebook-guided LLM event coding as a measurement problem rather than only a classification problem. Across predictive evaluations and codebook-change probes, we find that richer codebook guidance improves fine-grained event classification, but does not guarantee adherence to the coding rules under controlled codebook changes.

Our results show that enriched codebooks help LLMs apply definitions, examples, and boundary rules, especially for fine-grained event distinctions: mean root-level macro-F1 rises from 0.457 to 0.633. However, models still change their outputs when only the codebook order changes, and performance against the reassigned labels remains low.

This gap matters because codebook outputs often become data. If a model relies on surface cues rather than the intended coding rules, failures to follow the coding rules become uncertainty in the resulting measurements. Our findings therefore suggest that codebook-guided LLMs should be evaluated not only by predictive performance, but also by whether they preserve the coding rules under controlled changes to the codebook.

A longer-term direction is to build auditable coding systems that return more than a label: they should identify the codebook rule being applied, report uncertainty, flag cases that fail checks of rule adherence, and route them to human review.

\section*{Limitations}
\label{sec:limitations}

This study focuses on codebook-guided LLM measurement in political event coding, using a fine-grained PLOVER-based relation-classification task and a binary Cooperation/Conflict task. These settings are useful for testing codebook guidance and rule adherence, but they cover only part of the broader space of codebook-based text measurement, and political codebooks may themselves encode ontological assumptions or biases. Future work should examine whether the framework transfers to other codebooks and socio-political event coding settings, including protest coding, conflict event datasets, gun violence coding, policy coding, and incident classification \citep{hurriyetoglu2021case,barker2021case,radford2021case,pavlick2016gun}.

A second limitation concerns model coverage. We evaluate four open-source instruction-tuned models of 7 to 9 billion parameters and selected prior baselines, but larger frontier models and domain-adapted models may behave differently. The retrieval and chain-of-thought results should therefore be interpreted as findings about the implementations tested here rather than as general conclusions about these method families.

The method comparisons also do not provide full component-level ablations. The Enriched setting combines manually selected examples, event-mode guidance, boundary notes, and disambiguation rules, so the experiments do not isolate the contribution of each component. These results therefore compare practical codebook-guidance configurations rather than identify the causal effect of any single design choice.

The swapped-mapping probe is intentionally diagnostic: it tests whether models follow a supplied codebook when label semantics conflict with familiar meanings, but it does not cover all realistic ontology revisions. Its prompts also keep original label names in the instruction and in parts of the codebook, so its results cannot separate familiar label meanings from conflicting prompt guidance. We do not compare models with human coders under the same changes, and we do not measure how probe failures affect downstream analyses.

Future probes could also test actor-role perturbations, such as reversing source and target roles or evaluating the same sentence under different actor pairs. Finally, although we report bootstrap confidence intervals for the main PLV predictive comparisons, the codebook-change probes are reported as averaged point estimates without uncertainty estimates.

\section*{Ethical Considerations}

This work evaluates LLM-based political event coding as a measurement tool. Such systems should not be used to produce social-science data without auditing, uncertainty reporting, and human review for cases that fail checks of rule adherence. Because event-coded data may inform claims about conflict, cooperation, and political behavior, errors or failures to follow the coding rules can affect downstream analysis. This concern is consistent with broader discussions of LLMs in computational social science, where validity, reliability, and annotation quality remain central issues \citep{ziems2024can,he2025survey}.

We release the code, prompt variants, and codebooks used in our evaluations; model outputs are not included, and the PLV and AW data are available from their original sources. The codebook-change probes used here are intended as diagnostic tools for identifying such risks, not as a replacement for expert validation.

\section*{Acknowledgment}
U.S.~National Science Foundation support via 1) computational resources provided by the Chameleon testbed~\cite{keahey2020lessons} and 2) Brandt's research under Award OAC-2311142. 

\bibliography{references}

\clearpage

\appendix

\section{Prompt Templates and Worked Examples}
\label{app:prompts}

For the ZSP examples below, we use the following example: ``Indonesian students said they would stage mass demonstrations to demand political reforms by President Suharto's government.'' The source is Indonesian students, the target is President Suharto's government, and the gold label is THREATEN.

\subsection{ZSP Hypothesis Examples}
\label{app:zsp-examples}

ZSP Tiny constructs one hypothesis per Rootcode using only the label name:
\begin{quote}
{[S]} threatened {[T]}.\\
{[S]} protested against {[T]}.\\
{[S]} provided aid to {[T]}.
\end{quote}

For this ZSP example, REQUEST receives a higher entailment score than THREATEN in the flat pass.  This is because the flat pass has no mechanism for event-mode disambiguation.

ZSP Full replaces label-name hypotheses with codebook-description hypotheses. For THREATEN, the hypothesis states that the source made coercive or forceful warnings with serious potential repercussions directed at the target.

ZSP Tree processes the example in three levels. Level 1 scores Past-mode hypotheses and keeps REQUEST and PROTEST as top candidates. Level 2 branches surviving candidates into Past and Future variants; the future protest variant maps to THREATEN. Level 3 applies the class-disambiguation rule and produces the final prediction THREATEN.

\subsection{Prompting Conditions}
\label{app:prompting-conditions}

\paragraph{No Codebook.}
The prompt asks the model to classify the relation between the marked source and target and choose one label from the valid PLOVER label set. No definitions or examples are provided.

\paragraph{Compact.}
Compact corresponds to the compact codebook setting used in the main text. Each entry contains the root label, quad-level assignment, and a short natural-language definition. A global disambiguation rule prioritizes Material Conflict over Verbal Conflict when both are plausible.

\paragraph{Enriched.}
Enriched keeps the same label space as Compact but adds quad-level section headers, worked examples, event-mode guidance, and boundary-case disambiguation notes. The global rules address future-tense cooperation, negated or halted cooperation, peacekeeping forces, CONSULT overuse, and Material Conflict priority. For example, future cooperation is coded as AGREE, completed aid as AID, and halted aid as SANCTION.

\paragraph{Instruction and original label names.}
Each PLV codebook begins with an instruction. The Enriched instruction reads:

\begin{quote}
\small
Instruction: You are an expert annotator for political relation and event coding using the PLOVER 15-root ontology. Classify the relation between source (\texttt{<S></S>}) and target (\texttt{<T></T>}). Focus on the actual action described, not background context. Material Conflict (Q4) overrides Verbal Conflict (Q3)---e.g.\ ``protest to request'' = PROTEST; ``convict and arrest'' = COERCE. Future-tense cooperation = AGREE, not enacted material deployment. Negated or halted cooperation = SANCTION (e.g.\ halted military aid, cut diplomatic ties). Peacekeeping forces, workers, or observers sent to help = AID, not MOBILIZE. Use CONSULT only when meeting is the main action, not a byproduct of another action.\\
Task: single label multiclass\\
Output Reminder: Write the name of the Label that fits best, with no other text. For example, ``Label: AGREE'', ``Label: THREATEN'', or ``Label: ASSAULT''.
\end{quote}

The Compact instruction is shorter but also names PROTEST and COERCE in its priority rule and uses the same output examples. Some definitions also mention other labels by name; for example, the REQUEST definition states that demands made as demonstrations are coded as PROTEST. The swapped-mapping probe renames only the label of each codebook entry and leaves the instruction and these mentions unchanged. The generic-label probe replaces the output examples with \texttt{LABEL\_1} and \texttt{LABEL\_2} and leaves the other mentions unchanged.

\paragraph{CoT.}
CoT uses the Enriched codebook with an added step-by-step reasoning instruction before the final label prediction. It asks the model to reason through the source, target, main action, event mode, cooperation/conflict status, and final root label before answering.

\paragraph{Representative CoT prompt template.}
The following template shows the reasoning format used for CoT. The placeholder \texttt{<DOCUMENT>} is replaced with the input sentence containing the marked source and target spans.

\begin{quote}
\small
\textbf{System:} You are a helpful assistant. Follow the requested answer format.

\vspace{0.4em}
\textbf{User:} You are a political event classifier.

\vspace{0.4em}
\textbf{LABEL DEFINITIONS:}\\
This block contains the Enriched codebook entries: root label, definition, clarification, examples, and boundary guidance.

\vspace{0.4em}
\textbf{Sentence:} \texttt{<DOCUMENT>}

\vspace{0.4em}
\textbf{Think step by step:}\\
1. Who is source, who is target?\\
2. What is the main action?\\
3. Verbal (statements/promises) or material (physical)?\\
4. Cooperative or conflictual?\\
5. Which label fits best?

\vspace{0.4em}
After reasoning, write final answer as:

\textbf{ANSWER:} \texttt{<label>}

\vspace{0.4em}
\textbf{Allowed labels:} AGREE, CONSULT, SUPPORT, COOPERATE, AID, YIELD, REQUEST, ACCUSE, REJECT, THREATEN, PROTEST, SANCTION, MOBILIZE, COERCE, ASSAULT.
\end{quote}

\paragraph{ICL.}
The in-context learning prompt gives one labeled demonstration per root class as an input--output pair with a one-line reason. It does not include full codebook definitions.

\section{RAG Configuration Details}
\label{app:rag-details}

RAG uses all-MiniLM-L6-v2 sentence embeddings \citep{reimers2019sentence} and three FAISS indices: codebook definition chunks, disambiguation-rule chunks, and labeled examples. It retrieves input-relevant definitions and examples from the Enriched codebook instead of presenting the full codebook. Per-model macro-F1 is 0.665, 0.595, 0.547, and 0.545 on PLV (root) and 0.786, 0.771, 0.852, and 0.844 on AW (binary) for Gemma2, Qwen2.5, Mistral, and Llama3.1.

\section{Additional Codebook and Model Comparisons}
\label{app:additional-comparisons}

\subsection{Cross-Model PLV Predictive Checks}
\label{app:cross-model-plv}

\begin{table}[!t]
\centering
\caption{Cross-model PLV predictive checks. Scores are macro-F1 $\times$ 100 on the PLV test set. Model names are shortened for space. Across all four models, root-level performance improves from the no-codebook baseline to Compact and is highest under Enriched; CoT does not consistently improve performance.}
\label{tab:cross-model-plv}
\scriptsize
\setlength{\tabcolsep}{3pt}
\begin{tabular}{llrrr}
\toprule
Model & Method & Binary F1 & Quad F1 & Root F1 \\
\midrule
\multirow{5}{*}{Gemma2}
& No Codebook & 90.9 & 65.7 & 52.0 \\
& Compact & 92.7 & 80.4 & 63.6 \\
& CoT & 83.4 & 67.2 & 52.0 \\
& ICL & 94.7 & 71.5 & 57.3 \\
& Enriched & \textbf{96.0} & \textbf{86.9} & \textbf{71.8} \\
\midrule
\multirow{5}{*}{Qwen2.5}
& No Codebook & 88.2 & 62.9 & 50.4 \\
& Compact & 89.5 & 76.7 & 57.9 \\
& CoT & 79.2 & 65.3 & 47.9 \\
& ICL & 90.3 & 64.9 & 51.4 \\
& Enriched & 90.0 & 79.9 & 61.0 \\
\midrule
\multirow{5}{*}{Llama3.1}
& No Codebook & 86.8 & 55.9 & 39.5 \\
& Compact & 90.8 & 75.8 & 52.6 \\
& CoT & 84.2 & 70.4 & 50.3 \\
& ICL & 88.9 & 59.6 & 42.9 \\
& Enriched & 89.5 & 76.8 & 58.7 \\
\midrule
\multirow{5}{*}{Mistral}
& No Codebook & 91.3 & 54.4 & 40.9 \\
& Compact & 89.2 & 79.7 & 55.5 \\
& CoT & 75.9 & 60.6 & 40.5 \\
& ICL & 86.3 & 62.5 & 44.4 \\
& Enriched & 90.0 & 79.4 & 61.7 \\
\bottomrule
\end{tabular}
\end{table}

Table~\ref{tab:cross-model-plv} reports the full cross-model PLV predictive results, including binary, quad-level, and root-level macro-F1.

The cross-model results show the same qualitative pattern as the main PLV table. Root-level performance improves from the no-codebook baseline to Compact and is highest under Enriched for all four open-source models. CoT does not consistently improve over the no-codebook baseline and remains below Enriched at the root level for all four models.

\subsection{Bootstrap Confidence Intervals}
\label{app:bootstrap}

Table~\ref{tab:bootstrap-plv-delta} reports nonparametric bootstrap 95\% confidence intervals for the main PLV root-level macro-F1 differences. Compact improves over No Codebook for all four models, and Enriched improves over Compact for all four models; for Qwen2.5, the lower bound is close to zero and rounds to 0.000.

\begin{table}[!t]
\centering
\caption{Bootstrap uncertainty check for PLV root-level macro-F1 differences. The table reports method differences rather than absolute scores.}
\label{tab:bootstrap-plv-delta}
\footnotesize
\setlength{\tabcolsep}{3pt}
\resizebox{\columnwidth}{!}{
\begin{tabular}{llrr}
\toprule
Model & Comparison & $\Delta$ F1 & 95\% CI \\
\midrule
\multirow{2}{*}{Gemma2}
& Compact -- No Codebook & +0.116 & [0.079, 0.157] \\
& Enriched -- Compact    & +0.082 & [0.056, 0.107] \\
\addlinespace
\multirow{2}{*}{Qwen2.5}
& Compact -- No Codebook & +0.075 & [0.036, 0.108] \\
& Enriched -- Compact    & +0.031 & [0.000, 0.060] \\
\addlinespace
\multirow{2}{*}{Mistral}
& Compact -- No Codebook & +0.146 & [0.106, 0.184] \\
& Enriched -- Compact    & +0.062 & [0.034, 0.089] \\
\addlinespace
\multirow{2}{*}{Llama3.1}
& Compact -- No Codebook & +0.131 & [0.097, 0.166] \\
& Enriched -- Compact    & +0.061 & [0.026, 0.093] \\
\bottomrule
\end{tabular}
}
\end{table}

\subsection{Direct Binary and Quad-Level Classification}
\label{app:direct-bin-quad}

\begin{table}[!t]
\centering
\caption{Direct binary and quad-level macro-F1 $\times$ 100 using Gemma2:9B. GPT baselines are from \citet{hu2024leveraging}.}
\label{tab:direct-bin-quad}
\footnotesize
\begin{tabular}{lrr}
\toprule
Method & Binary F1 & Quad F1 \\
\midrule
Direct No Codebook & 90.5 & 64.3 \\
Direct Compact & 94.0 & 78.4 \\
Direct ICL & 90.5 & 76.2 \\
Direct CoT & 93.5 & 75.2 \\
GPT-4 baseline & 93.4 & 76.7 \\
GPT-3.5 baseline & 90.1 & 66.2 \\
\bottomrule
\end{tabular}
\end{table}

In this experiment, the model predicts the binary or quad label directly rather than a root label that is mapped upward. Direct binary and quad-level experiments show that codebook information is most useful when the task requires finer distinctions. Compact gives the highest binary and quad scores in this setting; CoT is close to Compact at the binary level and below it at the quad level.

\subsection{Full Probe Results}
\label{app:full-behavioral-probes}

The main text reports the PLV probe results. Tables~\ref{tab:plv-full-probes} and~\ref{tab:aw-full-probes} provide fuller probe-level details averaged across the four open-source models.

\paragraph{Metric computation.}
Orig. Acc. is original-condition probe accuracy, computed as the share of examples whose original-codebook prediction matches the gold label. Rev. Acc. and Shuf. Acc. are accuracies under reversed and shuffled codebook orders. Rev. Chg. and Shuf. Chg. measure the share of examples whose predictions change relative to the original-order prediction. Order $\kappa$ is Fleiss' $\kappa$ over the original, reversed, and shuffled prediction sets. Generic F1 is computed against the expected neutral label in the generic-label probe. Swap F1 is computed against the expected label under the perturbed label--definition mapping. Order statistics are not reported for RAG, which retrieves local chunks rather than presenting the full ordered codebook. Because AW has only two labels, its reversed and shuffled orders coincide.

\begin{table}[!t]
\centering
\caption{Full PLV probe results (312 examples) averaged across four open-source models. PLV is evaluated at the root-label level.}
\label{tab:plv-full-probes}
\scriptsize
\setlength{\tabcolsep}{3pt}
\renewcommand{\arraystretch}{0.92}
\resizebox{\columnwidth}{!}{%
\begin{tabular}{lrrrrrrrr}
\toprule
Method & Orig. & Rev. & Rev. Chg. & Shuf. & Shuf. Chg. & $\kappa$ & Gen. & Swap \\
\midrule
Compact  & 0.612 & 0.615 & 0.255 & 0.637 & 0.209 & 0.743 & 0.566 & 0.105 \\
Enriched & 0.619 & 0.625 & 0.212 & 0.640 & 0.188 & 0.782 & 0.519 & 0.081 \\
ICL      & 0.454 & 0.474 & 0.354 & 0.491 & 0.306 & 0.668 & 0.152 & 0.064 \\
CoT      & 0.642 & 0.640 & 0.256 & 0.667 & 0.236 & 0.745 & 0.625 & 0.152 \\
RAG      & 0.633 & N/A & N/A & N/A & N/A & N/A & 0.522 & 0.191 \\
\bottomrule
\end{tabular}%
}
\end{table}

\begin{table}[!t]
\centering
\caption{Full AW probe results (253 examples) averaged across four open-source models. AW is evaluated as a binary Cooperation/Conflict task.}
\label{tab:aw-full-probes}
\scriptsize
\setlength{\tabcolsep}{3pt}
\renewcommand{\arraystretch}{0.92}
\resizebox{\columnwidth}{!}{%
\begin{tabular}{lrrrrrrrr}
\toprule
Method & Orig. & Rev. & Rev. Chg. & Shuf. & Shuf. Chg. & $\kappa$ & Gen. & Swap \\
\midrule
Compact  & 0.869 & 0.854 & 0.153 & 0.854 & 0.153 & 0.755 & 0.859 & 0.141 \\
Enriched & 0.841 & 0.869 & 0.136 & 0.869 & 0.136 & 0.737 & 0.865 & 0.109 \\
ICL      & 0.807 & 0.824 & 0.083 & 0.824 & 0.083 & 0.854 & 0.029 & 0.148 \\
CoT      & 0.834 & 0.825 & 0.104 & 0.825 & 0.104 & 0.776 & 0.823 & 0.143 \\
RAG      & 0.837 & N/A & N/A & N/A & N/A & N/A & 0.009 & 0.137 \\
\bottomrule
\end{tabular}%
}
\end{table}

The fuller probe tables show the same pattern as the main text. In PLV, Enriched improves order agreement relative to Compact but does not improve Generic F1 or Swap F1. In AW, the binary label space yields stronger original-condition accuracy, and generic-label scores stay high for Compact, Enriched, and CoT (0.82 to 0.87) but fall near zero for ICL and RAG, and every method scores low under swapped mappings.

\subsection{Qualitative Example of a Swapped-Mapping Failure}
\label{app:qualitative-swapped}

Table~\ref{tab:qualitative-swapped-examples} shows a representative swapped-mapping case. Original denotes the gold label and original prediction. Swapped denotes the expected label under the perturbed label--definition mapping and the model prediction.

\begin{table}[!t]
\centering
\caption{A representative swapped-mapping failure on PLV. The swapped codebook names the threat definition AID, so a model following the codebook should output AID.}
\label{tab:qualitative-swapped-examples}
\scriptsize
\setlength{\tabcolsep}{2pt}
\renewcommand{\arraystretch}{1.08}
\begin{tabular}{p{0.18\columnwidth}p{0.74\columnwidth}}
\toprule
Method & Example \\
\midrule
RAG &
\textit{Text:} North Korea threatened to shell the Christmas tree. \newline
Original: THREATEN $\rightarrow$ THREATEN. \newline
Swapped: AID $\rightarrow$ THREATEN. \\
\bottomrule
\end{tabular}
\end{table}

\end{document}